\documentclass[a4paper,conference]{IEEEtran}
\usepackage{cite}

\ifCLASSINFOpdf
   \usepackage[pdftex]{graphicx}
\else
   \usepackage[dvips]{graphicx}
\fi
\usepackage{amsmath}
\usepackage{amsfonts}
\usepackage{algorithmic}

\usepackage{array}

\ifCLASSOPTIONcompsoc
 \usepackage[caption=false,font=normalsize,labelfont=sf,textfont=sf]{subfig}
\else
 \usepackage[caption=false,font=footnotesize]{subfig}
\fi
\usepackage{url}

\usepackage{multirow}
\usepackage{booktabs}
\usepackage{arydshln}

\begin{document}
%
\title{Interpretable Structured Learning with Sparse Gated Sequence Encoder for Protein-Protein Interaction Prediction}

\author{\IEEEauthorblockN{Kishan KC}
\IEEEauthorblockA{Golisano College of Computing and \\Information Sciences\\
Rochester Institute of Technology\\
Rochester, NY 14623\\
Email: kk3671@rit.edu}
\and
\IEEEauthorblockN{Feng Cui}
\IEEEauthorblockA{Thomas H. Gosnell School of \\Life Sciences\\ 
Rochester Institute of Technology\\
Rochester, NY 14623\\
Email: fxcsbi@rit.edu}
\and
\IEEEauthorblockN{Anne R. Haake}
\IEEEauthorblockA{Golisano College of Computing and \\Information Sciences\\
Rochester Institute of Technology\\
Rochester, NY 14623\\
Email: arhics@rit.edu}
\and
\IEEEauthorblockN{Rui Li}
\IEEEauthorblockA{Golisano College of Computing and \\Information Sciences\\
Rochester Institute of Technology\\
Rochester, NY 14623\\
Email: rxlics@rit.edu}}


%
\author{\IEEEauthorblockN{Kishan K C\IEEEauthorrefmark{1},
Feng Cui\IEEEauthorrefmark{2},
Anne R. Haake\IEEEauthorrefmark{1},
Rui Li\IEEEauthorrefmark{1}}
\IEEEauthorblockA{\IEEEauthorrefmark{1}Golisano College of Computing and Information Sciences,
Rochester Institute of Technology, NY, USA\\ Email: \{kk3671, arhics, rxlics\}@rit.edu}
\IEEEauthorblockA{\IEEEauthorrefmark{2}Thomas H. Gosnell School of Life Sciences, Rochester Institute of Technology, NY, USA \\
Email: fxcsbi@rit.edu}}


\maketitle

\begin{abstract}
Predicting protein-protein interactions (PPIs) by learning informative representations from amino acid sequences is a challenging yet important problem in biology. Although various deep learning models in Siamese architecture have been proposed to model PPIs from sequences, these methods are computationally expensive for a large number of PPIs due to the pairwise encoding process. Furthermore, these methods are difficult to interpret because of non-intuitive mappings from protein sequences to their sequence representation. To address these challenges, we present a novel deep framework to model and predict PPIs from sequence alone. Our model incorporates a bidirectional gated recurrent unit to learn sequence representations by leveraging contextualized and sequential information from sequences. We further employ a sparse regularization to model long-range dependencies between amino acids and to select important amino acids (protein motifs), thus enhancing interpretability. Besides, the novel design of the encoding process makes our model computationally efficient and scalable to an increasing number of interactions. Experimental results on up-to-date interaction datasets demonstrate that our model achieves superior performance compared to other state-of-the-art methods. Literature-based case studies illustrate the ability of our model to provide biological insights to interpret the predictions.
\end{abstract}

%
\IEEEpeerreviewmaketitle

\section{Introduction}
Proteins are the functional units within an organism that form molecular machines organized by their protein-protein interactions (PPIs) to carry out many biological and molecular processes. The study of PPIs not only plays a crucial role in understanding biological phenomena but also provides insights about the molecular etiology of diseases as well as the discovery of putative drug targets~\cite{martin2004predicting}. Although a large amount of reliable PPIs has been experimentally identified, the discovery of all possible PPIs through these experiments is intractable. Thus, efficient computational methods capable of accurately predicting PPIs and speeding up new PPI discovery are needed.

The primary structure of a protein, the protein sequence, determines the protein's unique three-dimensional shape, giving rise to an assumption that knowledge of the protein sequence alone might be sufficient to model the interaction between two proteins~\cite{anfinsen1973principles,shen2007predicting}. There is a longstanding interest in predicting PPIs from protein sequences which are by far the most abundant data available for proteins~\cite{yang2010prediction}. Traditional methods proposed to model PPIs involve extracting features based on domain expertise and training machine learning model on these features\cite{guo2008using, yang2010prediction, you2014prediction}. The performance of these methods relies heavily on the capability to select appropriate features, while the extracted features lack enough information about the interactions.
 
Recently, the state-of-the-art deep learning methods in the Siamese architecture have been proposed to automatically extract features from sequences to model interactions in an end-to-end framework~\cite{somaye:DPPI, chen2019pipr}. Siamese architecture for PPI prediction consists of two identical neural networks each taking one of the protein sequences and modeling their mutual influence. Although these methods perform quite well on small datasets, these methods involve the pairwise encoding process, making it computationally inefficient for a large number of interactions~\cite{hsu2015neural}. This has significant memory and runtime limitations when training with large datasets.

In this work, we propose a novel method to address these challenges. Specifically, we aim to learn a bidirectional GRU (BiGRU) model that maps variable-length sequences to a sequence of vector representations - one per amino acid position that encodes the sequential and contextual properties of amino acids. Since proteins interact with other proteins that perform different functions within a cell and even have different sequence patterns, we further encode the representation to the Gaussian distribution instead of a single point to capture uncertainty about its representation. We then define the cost function that incorporates the contrastive criteria between the latent Gaussian distributions such that the similarity between these distributions effectively captures complex interactions between proteins. It allows our model to minimize the statistical distance between interacting proteins while maximizing the distance for non-interacting proteins.

Alongside making accurate PPI predictions, it is crucial to have interpretable models that help domain experts understand how individual amino acids in the sequence contribute to the model's decisions. However, none of the state-of-the-art methods can provide such interpretability, limiting their practicality from biological perspectives. Since only a few amino acids in the interface region of the sequences are involved in interactions with other proteins~\cite{tonddast2015protein}, we, therefore, design a sparse and structured gate mechanism to guide the model to selectively focus on few amino acids in the sequence. The sparse gating mechanism outputs sparse weights - one per amino acid position - that explains how much contribution each amino acid makes and thus enhances interpretability.

Experimental results show that our method outperforms state-of-the-art methods on two challenging datasets: yeast and human PPIs from the BioGRID interaction database. Finally, we demonstrate that the learned sparse gate values correspond to the biologically interpretable protein motifs. A literature-based case study illustrates that our model effectively learns to identify the important residues from the sequence.

\section{Related work}
\label{related-work}
Traditional methods focus on extracting features from protein sequences such as autocovariance (AC)~\cite{guo2008using}, conjoint triads (CT)~\cite{shen2007predicting} and composition-transition-distribution (CTD)~\cite{yang2010prediction} descriptors and training a binary classifier on these features to predict PPIs~\cite{guo2008using,you2014prediction}. Since these extracted features only summarize the specific aspects of protein sequences such as physicochemical properties, frequencies of local patterns, and the positional distribution of amino acids, they lack enough information about the interactions.

Recently, deep learning architectures have been developed to address PPI prediction by automatically extracting useful features from protein sequences~\cite{somaye:DPPI, chen2019pipr}. These methods adopt deep-Siamese like neural networks to model the mutual influence between protein sequences. They use encoder based on a convolutional neural network (CNN) to capture local features and recurrent neural network (RNN) to capture sequential and contextualized features from protein sequences. The encoder encodes a pair of sequences to lower-dimensional sequence vectors and a binary classifier predicts the probability of interaction based on these sequence vectors.

Specifically, DPPI~\cite{somaye:DPPI} uses a deep-CNN based Siamese architecture that focuses on capturing local patterns from protein evolutionary profiles~\cite{hamp2015evolutionary}. However, it requires extensive effort in data-preprocessing, specifically in constructing evolutionary profiles from protein sequences using Position-Specific Iterative BLAST (PSI-BLAST)~\cite{altschul1997gapped}. To construct an evolutionary profile for a protein sequence, PSI-BLAST searches against NCBI non-redundant protein database with nearly 184 million sequences, which is time-consuming and makes it unscalable to a large number of protein sequences. However, PIPR~\cite{chen2019pipr} incorporates a deep residual recurrent convolutional neural network (RCNN) for PPI prediction using only the sequences of the protein pair. PIPR uses residual RCNN encoder that combines CNN to capture local features and RNN to capture sequential and contextualized features from protein sequences. The sequence representation for each protein is obtained by feeding its sequence through the deep sequence encoder with multiple layers of RCNN units. The non-intuitive mapping from protein sequences to their sequence representation makes the representation difficult to interpret.

\section{Method}
\label{section:framework}
A protein sequence $\mathbf{s}$ is a list of amino acids $[a_{1},a_{2},\ldots,a_{L}]$ where $a_{l}$ is the amino acid at position $l$ and L is the length of the sequence. We aim to map a protein sequence $\mathbf{s}$ to a lower-dimensional Gaussian distribution with formal format as follows: $f : \mathbf{s} \rightarrow (\mathbf{\mu}, \Sigma)$, where $\mu \in \mathbb{R}^d$, and $\Sigma \in \mathbb{R}^{d \times d}$ with $d \ll L$ denoting the dimension of the Gaussian distribution. We employ the Wasserstein distance as a measure of how different are the encoded Gaussian distributions of protein sequences. In this work, the goal of PPI prediction is to learn a model that predicts a smaller Wasserstein distance between the Gaussian distributions of interacting proteins and a larger difference for non-interacting proteins.

\begin{figure}[ht]
\centering
\includegraphics[width=0.8\linewidth]{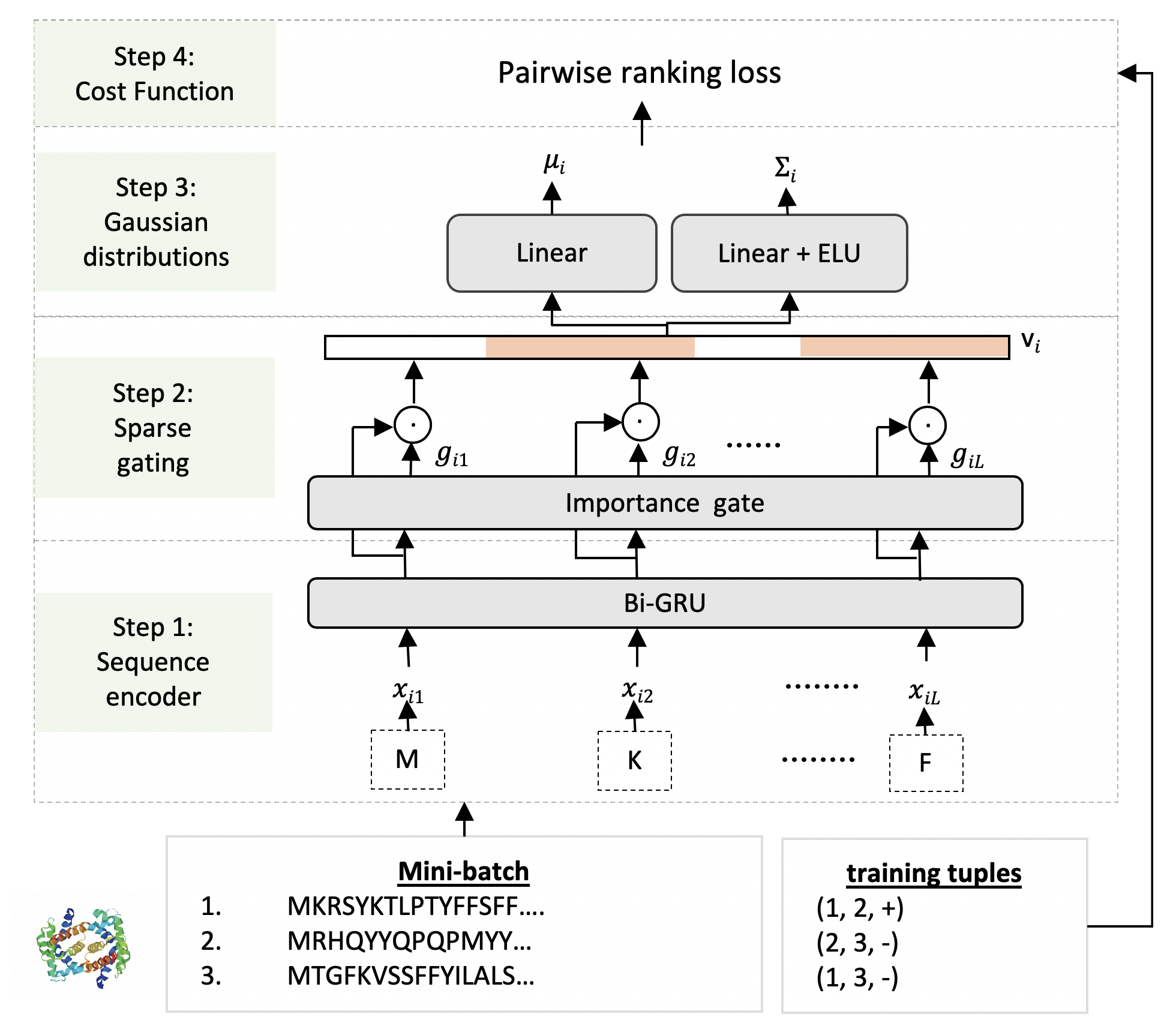}
\caption{Diagram of the model. A mini-batch of protein sequence is encoded to Gaussian distribution $\mu$ and $\Sigma$ using BiGRU encoder with sparse regularization. Model is trained by optimizing the pairwise Wasserstein distance between the training tuples (i.e. positive and negative interactions).}
\label{fig:architecture}
\end{figure}

We introduce our deep learning framework for PPI prediction from sequences. 
The overall architecture of the proposed framework is illustrated in Figure~\ref{fig:architecture}. In step 1, the sequence encoder incorporates a bidirectional gated recurrent unit (BiGRU) to encode the amino acid sequence to sequence of vector representations. In step 2, the importance gate models dependencies between all the positions in the sequence, which could allow it to directly model residue-residue dependencies. This enables the model to compute the importance of amino acid in each position based on the sequence of vector representations. In step 3, the representation of sequence is encoded into Gaussian representation with mean $\mu$ and $\Sigma$. In step 4, a pairwise ranking loss is employed to minimize the statistical distance between interacting proteins while maximizing the distance for non-interacting proteins.

\subsection{Sparse gating sequence encoder}
\subsubsection{\textbf{Protein sequence encoder}}
In step 1 as shown in Figure~\ref{fig:architecture}, the encoder takes a sequence of amino acids and encodes it to a sequence of vector representations - one per amino acid position. In particular, a one-hot encoded representation of amino acid $a_{l}$ is embedded to a vector representation $x_{l}$ through an embedding matrix: 
\begin{align}
x_{l} &= \textbf{W}_{e}a_{l}
\end{align}
where $\textbf{W}_e \in \mathbb{R}^{N \times d}$ is the weight of the embedding layer. 

To learn the sequential and contextualized representation of the amino acids in the sequence $\mathbf{s}$, we adopt the bidirectional Gated Recurrent Unit (BiGRU) that summarizes the sequence information from both directions. The sequence of vector representation $x_l$ is then fed into the bidirectional gated recurrent unit. It contains two encoding processes: a forward encoding $\overrightarrow{\operatorname{GRU}}$ which processes the sequence $\mathbf{s}$ from position 1 to $L$ , and a backward encoding $\overleftarrow{\operatorname{GRU}}$ which processes the sequence $\mathbf{s}$ from position $L$ to 1.

\begin{align}
h_{l} &= \operatorname{BiGRU}(x_{l}) = [\overrightarrow{\operatorname{GRU}}(x_{l}), \overleftarrow{\operatorname{GRU}}(x_{l})] 
\end{align}
where $h_{l}$ represents the GRU hidden state for amino acid $a_{l}$.  The representation of amino acid $h_l$ is the concatenation of forward hidden state $\overrightarrow{h}_l$ and backward hidden state  $\overleftarrow{h}_l$ which summarizes the information of whole sequence centered around $a_l$.
 
\subsubsection{\textbf{Sparse gating mechanism}}
Proteins bind to each other at specific binding domains on each protein. These domains can be just a few peptides long or span hundreds of amino acids. For this purpose, we introduce additional gates $\mathbf{g} = {\{g_1, g_2, \ldots g_L}\}$ indicating the activation of each amino acid. The amino acid $a_l$ is active if $g_l >$ 0 and is inactive when $g_l$ is 0. The gate values ${g}_{l}$ for the amino acid $a_{l}$, i.e., $g_{l} \in [0, 1]$ with 1 representing high importance. Let $S = \{l|g_l > 0\}$ be the set of amino acid that are active indicated by their respective gate values. We obtain the representation for protein sequences by scaling the hidden state $h_l$ with their respective gates:
\begin{align}
v_l &= {g}_{l} \odot h_{l} \\
\mathbf{v} &= \text{Concat}_{(l \in S)} (v_l)
\end{align}
where $\text{Concat}_{(l \in S)}$ represents the concatenation of sequence vectors for positions with $g_l> $ 0. $\odot$ denotes the elementwise product between gate values ${g}_{l}$ and GRU hidden state $h_{l}$. The sparse gate values $g_l$ leads to the sparse representation $\mathbf{v}$ of the sequence.  In contrast, for the positions with $g_l = 0$, the hidden states $h_l$ of these positions are not included in the representation $\textbf{v}$. The sparse gates act as the controllers to selectively activate the part of the network to account only for the subset of amino acids of the sequence.

We introduce an auxillary network that takes the GRU hidden states $h_{(.)}$ and generates the gate values for each position to determine whether the amino acid at that position is important for PPI prediction. The auxillary network models the long-range pairwise dependencies between amino acids in the sequence. With the auxillary network, our model explicitly considers dependencies between all position in the sequence, which could allow it to directly model residue-residue dependencies. In step 2 of Figure~\ref{fig:architecture}, GRU hidden states $h_{l}$ is transformed to score $p_l$ as:
\begin{align}
p_{l} &= \mathbf{W}_2(\tanh(\mathbf{W}_1h_{l}+\mathbf{b}_1))  + \mathbf{b}_2
\end{align}
where $\mathbf{W}_1 \in \mathbb{R}^{d \times d}, \mathbf{W}_2  \in \mathbb{R}^{d \times 1}$, $\mathbf{b}_1  \in \mathbb{R}^{d}$ and $\mathbf{b}_2  \in \mathbb{R}^{1}$ are the weight matrices and biases for the linear layers. Let $\mathbf{p} = [p_{1}, p_{2}, \ldots, p_{L}]$ is a vector of scores for amino acids in the sequence $\mathbf{s}$.  Next, we convert the vector of scores $\mathbf{p}$ to a probability distribution $\mathbf{g}$ so that $\sum_{l=1}^L g_l = 1$ and $0 \leq g_l \leq 1$. This allows us to quantify the relative contribution of each amino acid for the sequence representation. The softmax function is a simple choice to map the vector $\mathbf{p}$ to a probability distribution defined as:
\begin{align}
\text{softmax}_i(\mathbf{p}) = \frac{\exp(p_i)}{\sum_j\exp(p_j)}
\end{align}
Since the resulting softmax distribution has full support, i.e., $\text{softmax}(p_{l}) > 0$  for every $a_{l}$, it forces all the amino acids in the sequence to receive some probability mass. Sotmax distribution assigns weights to each amino acid and even an unimportant amino acids have small weights, the weights on important amino acids become much smaller for long sequences, leading to degraded performance. However, not all of the amino acids in the sequence contribute towards the certain functions or interactions. 

We introduce a sparsity regularization on $\mathbf{p}$ to only select the important set of amino acids, which may corresponds to the interface and may be important for interaction prediction. We employ sparsemax~\cite{martins2016softmax} as gate mechanism:
\begin{align}
\label{sparsemax}
\text{sparsemax}(\mathbf{p}) := \operatorname*{argmin}_{\mathbf{g} \in \Delta^K} \| \mathbf{g} - \mathbf{p}\|^2,
\end{align}
where $\Delta^K := \{ \in \mathbb{R}^K |\hspace{1pt} \mathbf{g} \geq 0, \mathbf{1}^T \mathbf{g} = 1 \}$ and $\mathbf{g}$ represents the Euclidean projection of $\mathbf{p}$ onto the $(K-1)$-dimensional probability simplex. The projection is likely to hit the boundary of the simplex, leading to sparse outputs, which allows the encoder to select only an informative subset of amino acids in the sequence. Although sparsemax leads to sparse representation, the sparse weights may only capture few important amino acids but may not identify a relatively long stretches of amino acids.  

Furthermore, to enable our model to selectively focus on relatively longer stretches of amino acids, we present fusedmax regularization~\cite{niculae2017regularized} that not only results in the sparse representations but also encourages the encoder to assign equal weights for contiguous sets of amino acids of the sequence while predicting interactions:

\begin{align}
\label{structuredattention}
\text{fusedmax}(\mathbf{p}) := \operatorname*{argmin}_{\mathbf{g} \in \Delta^K} \frac{1}{2}\|\mathbf{g} - \frac{\mathbf{p}}{\gamma}\|^2 + \lambda \sum_{j=1}^{L-1} |\mathbf{g}_{j+1} - \mathbf{g}_j|
\end{align}
where $\gamma$ controls the regularization strength and $\lambda$ is the tuning parameter that balances the attention to produce sparse outputs and to assign equal weights to the amino acids within each segment. In Eq. \ref{structuredattention}, the first part projects the scores $\mathbf{p}$ to the probability simplex and the second part encourages paying equal attention to adjacent amino acids in the sequence.  This allows the model to identify long stretches of amino acids that may bind with the residues from the interacting proteins. 

\subsubsection{\textbf{Encoding sequences as Gaussian distributions}}
Within an organism, a given protein may be involved in a complex interplay with various proteins that perform different functions within a cell and even have different sequence patterns. Such differences should be reflected in the uncertainty of its representation $\mathbf{v}$. To model the uncertainty about the representation~\cite{bojchevski2018deep,zhu:dvne}, sequence representation $\mathbf{v}$ is then encoded to $\mu$ and $\Sigma$ in the final layer of the architecture as shown in step 3 in Figure~\ref{fig:architecture}. To ensure that covariance matrices $\Sigma$ is positive definite, we use Exponential Linear Unit~\cite{clevert2015fast} in the final layer.
\begin{align}
    \mu & = \mathbf{W}_{\mu} \mathbf{v} + \mathbf{b}_{\mu} \label{mu} \hfill \\
    \Sigma & = \text{ELU}(\mathbf{W}_{\Sigma} \mathbf{v} + \mathbf{b}_{\Sigma}) + 1 \label{sigma}
\end{align}
where $\mathbf{W}_{\mu}, \mathbf{W}_{\Sigma}, \mathbf{b}_{\mu} $ and $\mathbf{b}_{\Sigma}$ denote the weight matrices and biases of linear layers that project intermediate representation $\mathbf{v}$ to mean $\mu$ and variance $\Sigma$ of Gaussian representation of the sequence $\mathbf{s}$. We denote a protein sequence $\mathbf{s}$ with a $d$-dimensional Gaussian distribution ($\mu, \Sigma$), where $\mu$ represents the center point of the sequence representation in the latent space and $\Sigma$ represents the uncertainty associated with the representation. With an assumption that the different dimensions of the latent Gaussian distribution are independent of each other, the covariance matrix $\Sigma$ is a diagonal matrix.
 
\subsection{Loss function definition} We employ the Wasserstein distance to measure the similarity between the Gaussian distributions of the proteins to make PPI prediction. Wasserstein distance allows the model to capture the transitivity property of PPIs that measures the tendency of proteins to cluster together into functional modules and protein complexes~\cite{liu2019fusing}.

The $p^{th}$ Wasserstein distance between two probability measures $\mu$ and $\nu$ is defined as:
\begin{align}
     W_p(\mu, \nu)^p &=  \inf \mathbb{E}[d(X, Y)^p] 
     \label{pth}
\end{align}
where $\mathbb{E}[Z]$ denotes the expected value of a random variable $Z$ and the infimum is taken over all joint distributions of random variables X and Y with marginals $\mu$ and $\nu$ respectively. Wasserstein distance is a well-defined measure that preserves both the symmetry and triangular inequality~\cite{Givens:w2}.

Wasserstein distance has a closed-form solution for two multivariate Gaussian distributions. This allows us to employ $2$-Wasserstein distance (abbreviated as $W_2$) as similarity measure between the latent Gaussian distributions $\mathcal{N}(\mu_i, \Sigma_i)$ and $\mathcal{N}(\mu_j, \Sigma_j)$: 
\begin{align}
dist &= W_2(\mathcal{N}(\mu_i, \Sigma_i), \mathcal{N}(\mu_j, \Sigma_j)) \label{dist}\\
    dist^2 &= ||\mu_i-\mu_j||_2^2 + \text{Tr}(\Sigma_i + \Sigma_j - 2(\Sigma_i^{\frac{1}{2}}\Sigma_j\Sigma_i^{\frac{1}{2}})^{\frac{1}{2}})\label{W2}
\end{align}

Since we focus on diagonal covariance matrices, thus $\Sigma_i\Sigma_j = \Sigma_j\Sigma_i$:
\begin{align}
    dist^2 = ||\mu_i-\mu_j||_2^2 + ||\Sigma_i^{\frac{1}{2}}-\Sigma_j^{\frac{1}{2}}||_F^2
\end{align}

According to the above equation, the time complexity to compute the $W_2$ between two multivariate Gaussian distributions is linear with the embedding dimension $d$. Since the computation of $W_2$ no longer constitutes a computational challenge, we choose $W_2$ as a measure of distance.

We use a pairwise ranking formulation with respect to the Wasserstein distance $W_2$ to model PPIs:
\begin{align}
\label{pairwise_w2}
    W_2(\mathcal{N}(\mu_i, \Sigma_i), \mathcal{N}(\mu_j, \Sigma_j)) < W_2(\mathcal{N}(\mu_i, \Sigma_i), \mathcal{N}(\mu_k, \Sigma_k))
\end{align}
where $\mathcal{N}(\mu_i, \Sigma_i)$ is the latent Gaussian distribution of sequence $\mathbf{s}_i$, and $(\mathbf{s}_i,\mathbf{s}_j)$ and $(\mathbf{s}_i,\mathbf{s}_k)$ represents positive and negative interaction respectively. Specifically, the idea of ranking formulation is to penalize ranking errors based on the the Wasserstein distances between the pairs.
The smaller the Wasserstein distance, the larger the possibility of interactions.

Finally, we employ square-exponential loss~\cite{LeCun06atutorial} to enable learning from the known pairwise interactions. Mathematically, the energy between the protein pairs can be defined as $E_{ij}$ = $W_2(\mathcal{N}(\mu_i, \Sigma_i),\mathcal{N}(\mu_j, \Sigma_j))$. Then, the loss function to be optimized is:
\begin{align}
\mathcal{L} &=  \sum_i \sum_{(i,j) \in \textbf{Y}^{+}} \sum_{(i, k) \in \textbf{Y}^{-}}
({E_{ij}}^2 + \exp(-E_{ik}))
\label{loss}
\end{align}
where $\textbf{Y}^+$ represents set of positive interactions and  $\textbf{Y}^-$ represents set of negative interactions. In our setting, the objective function penalizes the pairwise errors by the energy of the pairs, such that the energy of positive interactions is lower than the energy of negative interactions. Furthermore, such ranking formulation also maximizes the difference in energy between positive pair ($i, j$) and negative pair ($i, k$). Equivalently, this will make the possibility of interactions between the interacting proteins larger than that of non-interacting proteins.

Finally, we can optimize the parameters $\Theta$ (i.e. weights and biases) of the model such that the loss $\mathcal{L}$ is minimized and the pairwise rankings are satisfied. Specifically, for each protein, the distance with interacting proteins should be smaller than with non-interacting proteins. We term this as the ranking approach since interacting proteins have smaller $W_2$ distance and are ranked higher than non-interacting proteins.

\subsection{PPI prediction}
The Wasserstein distances between the latent Gaussian distributions of protein sequences corresponds to the possibility of their interaction. However, predicting PPIs by only computing the Wasserstein distance fails to take into account the homodimers, the proteins with identical sequences~\cite{somaye:DPPI}. The encoded Gaussian representations of these protein sequences will be the same and their Wasserstein distance will be 0 indicating they must interact. 

To overcome this limitation, we define pairwise features for all protein pairs by the concatenation of the absolute element-wise differences of means and variances and the element-wise multiplications of the means of their Gaussian representations, $ [|\mu_i - \mu_j|;|\Sigma_i - \Sigma_j|;\mu_i \odot \mu_j]$. This featurization is effective in modeling the symmetric relationship between proteins.
To predict binary interactions, we train a binary classifier on these pairwise features to learn a decision boundary $\delta$ that separates interacting proteins from non-interacting pairs.

\subsection{Efficient training}\label{training_time}
Siamese networks are suitable to train with contrastive loss mentioned in Eq.~\ref{loss}. However, Siamese training is inefficient when the amount of PPIs increases. In particular, the possible number of interactions for $N$ proteins is $(N^2 + N)/2$ (including self-interactions), which is computationally intensive for Siamese training. A mini-batch of $m$ interactions in Siamese training may have multiple occurrences of the same proteins, leading the sequence to be feed-forwarded for each interaction. It is sufficient to feed-forward each sequence once to compute the loss. To address this problem, we encode the minibatch of $n$ protein sequences and retrieve positive and negative interactions that involve them to compute the loss in the minibatch. With this setting, we are only required to feed-forward $N$ protein sequences compared to $(N^2 + N)/2$  pairs in the Siamese setting, which makes our method computationally efficient and scalable to a large number of interactions.

\section{Experiments}
\label{section:experiments}
We evaluate our method on two real-world datasets: yeast and human proteins to predict their interactions. We use the area under the ROC curve (AUROC) and the average precision (AP) scores as the evaluation metrics. With these evaluation metrics, we expect the positive protein pair to have higher interaction probability compared to negative protein pair.

\subsection{Experimental Setup}
\subsubsection{\textbf{Datasets}} The datasets for protein sequences of yeast and human proteins are from the EMBL-EBI Reference Proteome~\cite{dessimoz2012toward}. The information about the subcellular localization of proteins is extracted from UniProt database~\cite{uniprot2018uniprot}. The evolutional protein profiles for yeast and human protein sequences are collected from Rost Lab~\cite{hamp2015more}. We evaluate our proposed model with two types of protein features: (a) amino acid sequences and (b) evolutionary protein profiles constructed from these sequences.

To evaluate the performance of deep learning models, the interaction datasets are downloaded from the up-to-date BioGRID interaction database (Release 3.5.169)~\cite{oughtred2018biogrid}. The BioGRID database provides a large number of PPIs allowing us to evaluate the scalability of different approaches as well. Only the interactions that correspond to the physical binding between the protein pairs (say $\textbf{Y}^+$) are considered since these interactions are supported by experimental evidence. The negative samples $\textbf{Y}^-$ are generated by randomly sampling from all protein sequence pairs ($(\mathbf{s}_i, \mathbf{s}_j) \not\in \textbf{Y}^+$), that are not yet confirmed by experimental evidence. Furthermore, these negative interactions are filtered based on their subcellular localization, assuming that proteins in the different locations are unlikely to interact although some proteins do translocate~\cite{hamp2015more}.

Also, we perform the cluster analysis with the CD-HIT~\cite{li2006cd} to cluster protein sequences based on a certain similarity threshold that represents sequence identity. We remove the interactions such that no two proteins have a pairwise sequence identity greater than 10\%. Table~\ref{tab:statistics} shows the statistics of the interaction datasets.
\begin{table}[ht]
\centering
\caption{Statistics of interaction datasets}\label{tab:statistics}
\begin{tabular}{cccc}  
\toprule
\multirow{2}{*}{Data} &  No. of  & No. of  & No. of  \\
& proteins &  positive pairs & negative pairs \\
\midrule
Yeast & 3,651 & 50,344 & 50,376\\
Human & 7,028 & 73,624 & 73,628\\
\bottomrule
\end{tabular}
\end{table}

\begin{table*}[ht]
\centering
\caption{Average AUROC and AP scores (with standard deviation) averaged over five independent runs for PPI prediction. * represents statistically significant differences with PIPR (P-value $<$ 0.005). \label{tab:performance}
}
\begin{tabular}{lcccccc} 
\toprule
\multicolumn{2}{c}{\multirow{2}{*}{Method}} & \multirow{2}{*}{Data}  & \multicolumn{2}{c}{Yeast} & \multicolumn{2}{c}{Human} \\
&&&AUROC&AP&AUROC&AP \\
\midrule
\multicolumn{2}{l}{DPPI~\cite{somaye:DPPI}}& Profiles & 0.891$\pm$0.004  & 0.857$\pm$0.007  & 0.870$\pm$0.004  & 0.835$\pm$0.005  \\
\multicolumn{2}{l}{PIPR~\cite{chen2019pipr}} & sequences & 0.909$\pm$0.003 & 0.912$\pm$0.004 & 0.878$\pm$0.002  & 0.882$\pm$0.003\\
\hdashline
\multirow{4}{*}{\textbf{Our method (sparsemax)}} & \multirow{2}{*}{Ranking} & Profiles & 0.882$\pm$0.003 & 0.888$\pm$0.002 & 0.884$\pm$0.003 & 0.893$\pm$0.004\\
& & Sequences &  0.901$\pm$0.002 & 0.904$\pm$0.002 &
0.881$\pm$0.002 & 0.889$\pm$0.001  \\
& \multirow{2}{*}{Random Forest}& Profiles & 0.908$\pm$0.002 & 0.913$\pm$0.003 & \textbf{0.891 $\pm$0.005$^*$} & \textbf{0.896$\pm$0.005$^*$}\\
& & Sequences & \textbf{0.924$\pm$0.002$^*$} & \textbf{0.925$\pm$0.001$^*$} & 0.887$\pm$0.002 & 0.894$\pm$0.001 \\
\hdashline
\multirow{4}{*}{\textbf{Our method (fusedmax)}} & \multirow{2}{*}{Ranking} & Profiles & 0.882$\pm$0.006 & 0.885$\pm$0.006 & 0.873$\pm$0.09&0.881$\pm$0.01 \\
& & Sequences & 0.898$\pm$0.001 & 0.900$\pm$0.002 & 0.874$\pm$0.002& 0.883$\pm$0.001  \\
& \multirow{2}{*}{Random Forest}& Profiles & 0.906$\pm$0.004 & 0.912$\pm$0.005 & 0.872$\pm$0.015 & 0.877$\pm$0.015\\
& & Sequences & 0.919$\pm$0.003  & 0.921$\pm$0.002 & 0.881$\pm$0.002 & 0.886$\pm$0.001  \\
\bottomrule
\end{tabular}
\end{table*}

\subsubsection{\textbf{Hyperparameters and training details}}\label{implementation}
For both datasets, we train a sequence encoder with the same configuration. The best hyperparameters of our model are selected based on validation performance. The maximum length of the input sequence to the encoder is 1024 for efficient training. In the datasets, 91.2\% of yeast sequences and 86\% of human sequences are shorter than 1024 residues. 

The encoder consists of a BiGRU layer with 16 hidden units each, to map a protein sequence to a sequence of 32-dimensional representation, one per amino acid. Then, this representation is encoded to a latent Gaussian distribution with dimension $\mathbf{d}$ = 256 ($\mathbf{d} = 2 \times d$  = 128 for the mean and 128 for the variance of the Gaussian). 

All the weight matrices of the encoder layer are initialized using Xavier initialization~\cite{glorot2010understanding}. The model is trained on a single NVIDIA GeForce RTX 2080 Ti GPU for all experiments using Adam optimizer~\cite{kingma2014adam} with learning rate 0.003 and other default parameters provided by PyTorch~\cite{paszke2017automatic}. Our unique approach of encoding unique sequences allows us to train the model efficiently even with large batch sizes. We empirically find that our model converges in a small number of iterations ($\leq$ 50 for all shown experiments).

\subsection{Results on PPI prediction}
We compare our method against the state-of-the-art deep learning methods on the up-to-date BioGRID interaction datasets. We split the interactions into training, validation, and test sets (0.6:0.2:0.2). All the models are trained on the same training set and the best set of hyperparameters are selected based on their performances on the validation set. Finally, the models are evaluated on independent test sets. Table~\ref{tab:performance} reports the mean AUROC and AP and their standard errors on five independent runs. We perform a two-tailed Welch's t-test and Benjamini-Hochberg procedure to adjust p-value and find that the improvement over PIPR, the state-of-the-art method is statistically significant.

With the ranking approach, we expect our model to rank positive interactions higher than negative interactions, i.e. the probability of interactions between interacting protein pairs is greater than that of non-interacting protein pairs. Table~\ref{tab:performance} demonstrates that our model ranks positive interactions higher than negative interactions. The ranking based model with sparsemax regularization achieves comparable performance with PIPR. Furthermore, the random forest classifier trained to account for homodimeric interactions improve the model's performance on both datasets. The best parameters for random forest classifier are selected via grid search.

Furthermore, we evaluate our proposed method on evolutionary protein profiles. Protein profiles constructed from protein sequences capture the correlation between different proteins as well as between different parts of the sequences~\cite{cong2019protein}. Our model trained with protein profiles outperforms DPPI, the state-of-the-art deep architecture that uses profiles as the input to their deep Siamese model. Table~\ref{tab:performance} shows that our model with sequences achieves comparable or better performance compared to profiles across both datasets. This demonstrates that our model is capable of extracting useful information about interactions from protein sequences and alleviates the expensive process of profile construction from sequences.

\subsection{Ablation study of framework components}\label{ablation_experiments}
We next evaluate individual model components on the PPI prediction task with yeast dataset.

\subsubsection{\textbf{{Gaussian representations outperform point representations}}}
We first explore whether the Gaussian representation of sequences improves the performance of the model over point representation. We encode intermediate representation $\mathbf{v}$ of sequence $\mathbf{s}$ to point representation $\mathbf{z} = \mathbf{W}_\mathbf{z}\mathbf{v} + \mathbf{b}_\mathbf{z}$ instead of Gaussian representation (as in Eq.~\ref{mu} and \ref{sigma}) and define L2 norm as the similarity between the point representations of sequences $\mathbf{s}_i$ and $\mathbf{s}_j$ instead of Wasserstein distance (Eq. \ref{dist}):
\begin{align}
    dist^2 = ||\mathbf{z}_i-\mathbf{z}_j||_2^2
\end{align}
where $\mathbf{z}$ represents point representation of sequences $s$. Table~\ref{tab:ablation} shows that Gaussian representations are better predictors of PPIs compared to point representations. 

\begin{table}[ht]
\centering
\caption{Study of individual model components on Yeast dataset. The model trained without any gate mechanism (No gating), with different representations (point vs Gaussian) coupled with different regularization and random forest (RF) classifier.\label{tab:ablation}}
\begin{tabular}{cccc}  
\toprule
\multicolumn{2}{c}{Model configuration} &AUROC&AP \\
\midrule
\multicolumn{2}{c}{No gating} &0.880$\pm$0.001 & 0.875$\pm$0.003\\
\hdashline
\multirow{3}{*}{Point + RF} & {Softmax} & 0.881$\pm$0.001 & 0.877$\pm$0.001\\
&Fusedmax & 0.909$\pm$0.001 & 0.912$\pm$0.002\\
&Sparsemax & 0.913$\pm$0.001 & 0.916$\pm$0.002 \\
\hdashline
\multirow{3}{*}{Gaussian + RF} &Softmax & 0.882$\pm$0.001 & 0.879$\pm$0.002\\
 &{Fusedmax} & 0.919$\pm$0.003  & 0.921$\pm$0.001 \\
&{Sparsemax} & \textbf{0.924$\pm$0.002} & \textbf{0.925$\pm$0.001} \\
\bottomrule
\end{tabular}
\end{table}

\subsubsection{\textbf{\emph{Sparse gating mechanism improves performance}}}
We further demonstrate the importance of the proposed sparse gating mechanism by comparing the performance of the sequence encoder trained with and without various gating mechanism. We train models with different settings of the gate mechanism. Table~\ref{tab:ablation} demonstrates that the sparse gating mechanism provides a significant improvement over no gate mechanism and softmax in PPI prediction. This indicates that sparse regularization helps the model to selectively activate the important stretches of amino acids that are important to model and predict PPIs.

\subsubsection{\textbf{\emph{Dimension of Gaussian distribution is important}}}
\begin{figure}[!ht]
    \subfloat[]{\label{fig:a}\includegraphics[width=0.48\linewidth]{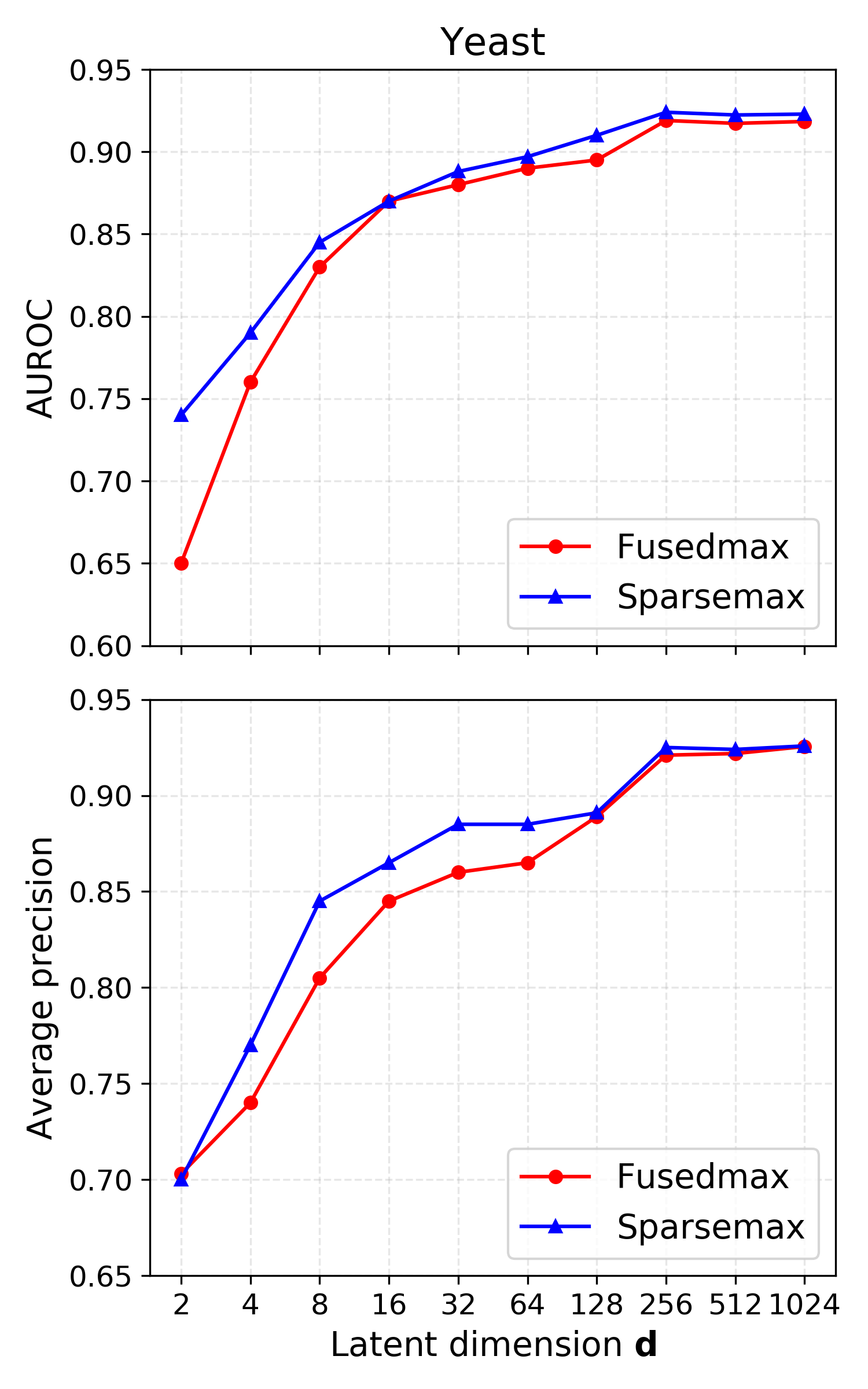}}
    \subfloat[]{\label{fig:b}\includegraphics[width=0.48\linewidth]{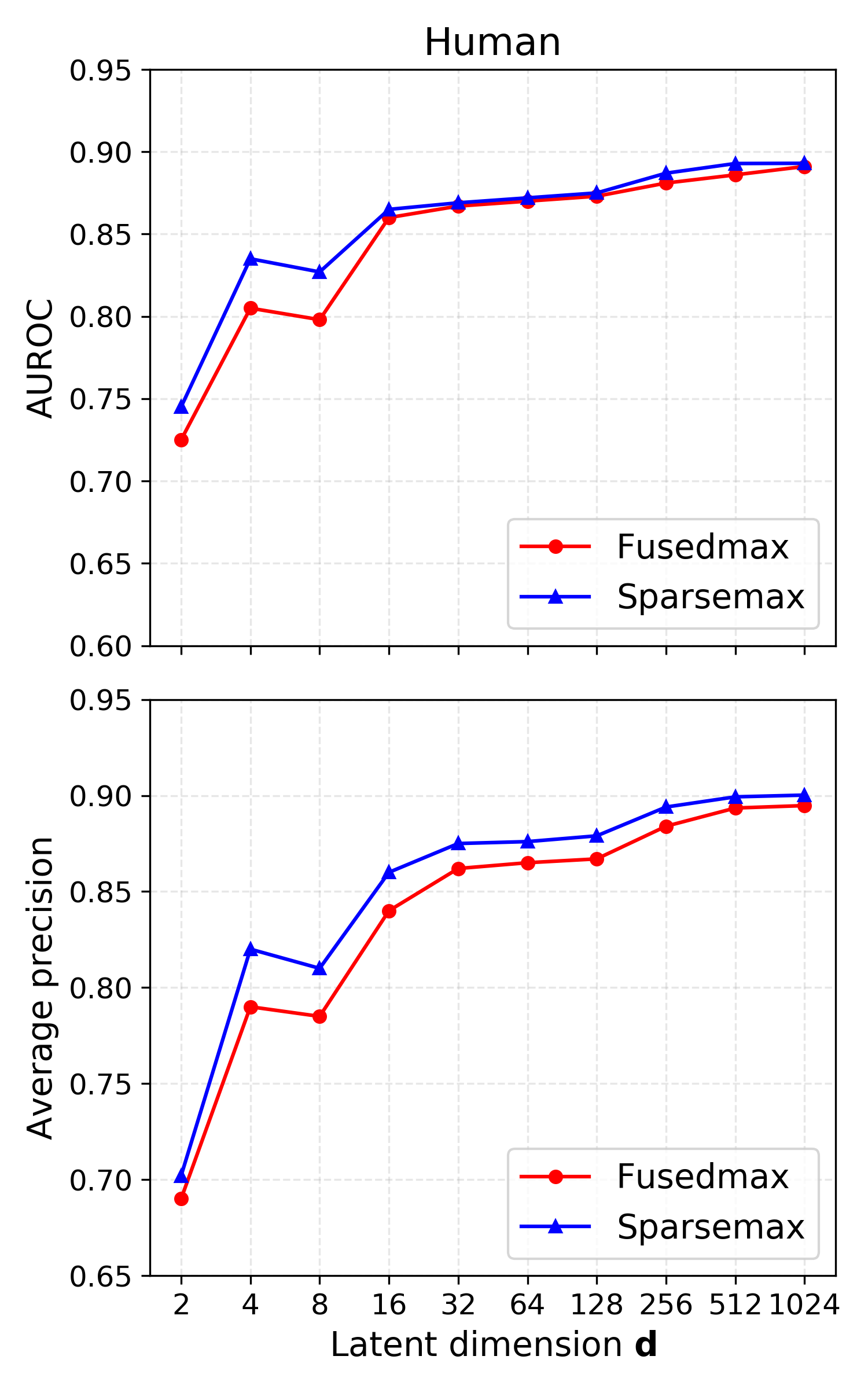}} 
    \hfill
    \raggedleft \caption{Effect of dimension of latent Gaussian distributions on the model's performance with different sparse gating mechanism for (a) yeast (b) human.}
    \label{fig:ablation}
\end{figure}

Finally, we investigate how the dimensionality of the latent Gaussian distributions can affect the model's performance.  Figure~\ref{fig:ablation} shows the plot of the AUROC and the AUPR scores of our method across the two organisms. When the dimension $\mathbf{d} (= 2 \times d)$ of the Gaussian distribution increases from $2$ to $256$, the performance also increases. When $\mathbf{d} \geq 128$, two regularization strategies result in similar performance. Moreover, the performance also remains stable when $\mathbf{d} \geq 256$.

\subsection{{Training time per epoch}} 
\begin{figure}[!ht]
\centering
\includegraphics[width=0.6\linewidth]{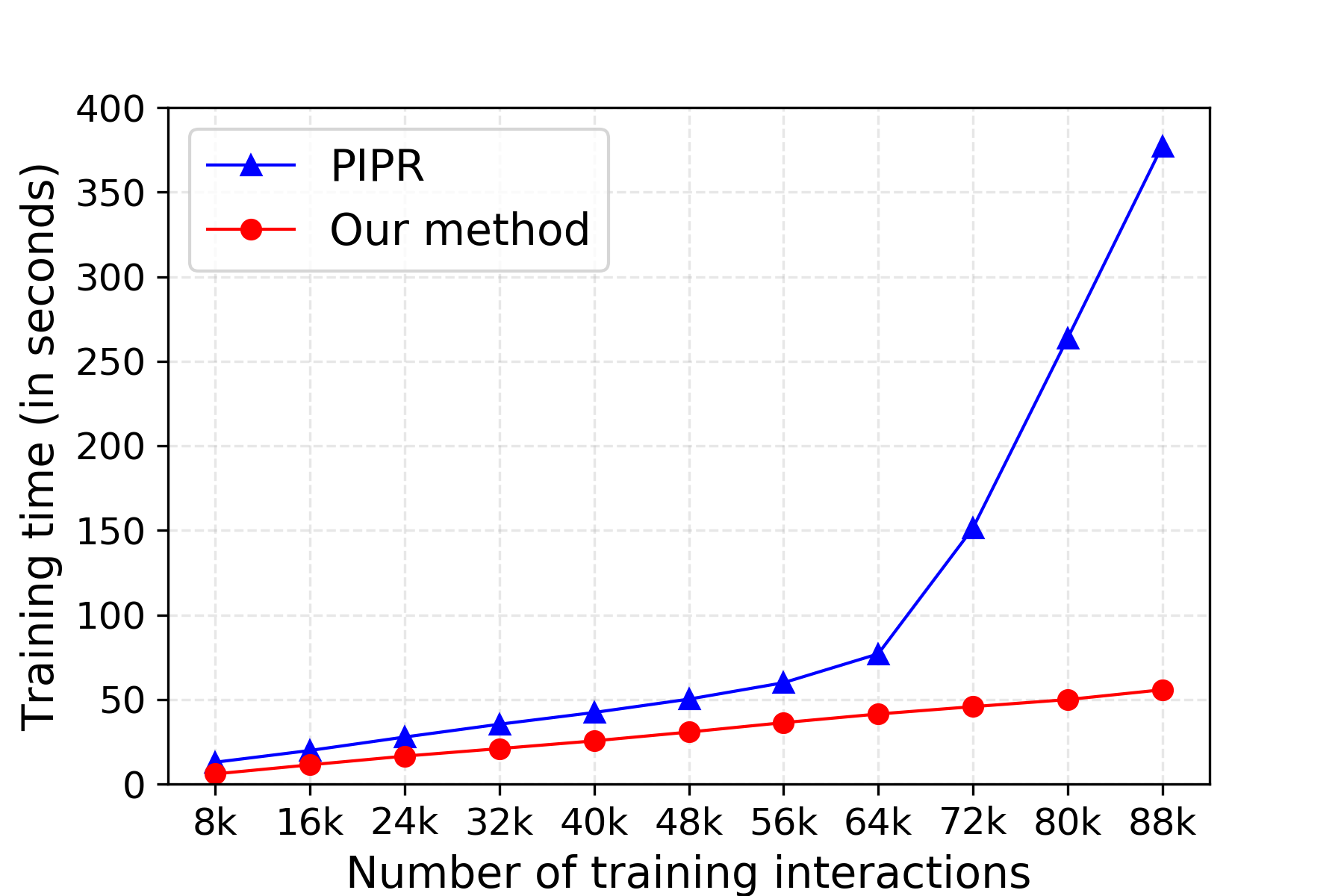}
\caption{Average training time per epoch. Our method is much faster than PIPR. Encoding the unique set of protein sequences to their latent Gaussian distribution and optimizing the loss based on their interactions makes our model efficient.}
\label{fig:epoch}
\end{figure}

Here, we compare the training time of our method and PIPR, the state-of-the-art deep learning model~\cite{chen2019pipr}. For fair comparison, we train both models in the same machine on the same dataset and compare only the average training time per epoch in Figure~\ref{fig:epoch}.  For this experiment, we randomly sample 8k, 16k, 24k, 32k, 40k, 48k, 56k, 64k, 72k, 80k, and 88k training interactions. 

Figure~\ref{fig:epoch} shows that our method is efficient in comparison to PIPR. PIPR uses a pairwise training process that requires a higher number of matrix multiplications for each interaction. On the other hand, given the large batch of interactions, our model finds the unique set of protein sequences involved in these interactions and encodes them, which significantly reduces the number of matrix multiplication. Once we have the embeddings for these sequences, we can compute the loss based on their respective interactions. As discussed in Section~\ref{training_time}, for instance, if there are 1000 interactions in a batch with 100 unique proteins, our model encodes only 100 protein sequences instead of 1000 protein pairs as in PIPR. Note that this approach allows our model to train on a large batch of interactions, and thus takes less time to train. This demonstrates that our method scales with the number of interactions and is significantly faster than PIPR.

\subsection{Interpretability}\label{qualitative}
Since our proposed model selectively activates the part of a given sequence, it is important to evaluate whether the selected parts are important. To explore this, we perform the quantitative evaluation on how the amino acids selected by the sparse gating mechanism in our proposed method align with the motifs from the Pfam motif library~\cite{finn2014pfam} from GenomeNet~\cite{kanehisa1997linking}. The gate vector $\mathbf{g}$ for a sequence $\mathbf{s}$ helps us interpret how much contribution an amino acid on that position signaled by GRU hidden state makes. Since our model computes gate values between 0 and 1 for each amino acid, we consider amino acids with $g_l > 0$ to be active i.e. used by the model for the representation of proteins. Table~\ref{motif_alignment} shows the average percentage of amino acids selected by the sparse gating mechanism and their alignment with motifs having biological significance. For instance, for yeast dataset, only 19.24\% amino acids (on average) are selected with fusedmax and 59.05\% of these selected amino acids aligns with the motif. This illustrates that the amino acids in the sequence selectively activated by our model to learn protein representation align with biologically interpretable motifs.

\begin{table}[ht]
\centering
\caption{Comparison of selected amino acids with the motifs from Pfam motif database}
\begin{tabular}{cccc}  
\toprule
\multirow{2}{*}{Dataset}&\multirow{2}{*}{Gating}& Selected & Alignment \\
&& amino acids ($\%$) & with motifs ($\%$) \\
\midrule
\multirow{2}{*}{Yeast}&Sparsemax&8.06& 49.96\\
&Fusedmax&19.24& 59.05\\
\hdashline
\multirow{2}{*}{Human}&Sparsemax&9.15& 48.33\\
&Fusedmax&23.33& 65.63\\
\bottomrule
\end{tabular}
\label{motif_alignment}
\end{table}

\begin{figure}[ht]
\centering     
\subfloat[]{\label{fig:int_a}\includegraphics[width=\linewidth]{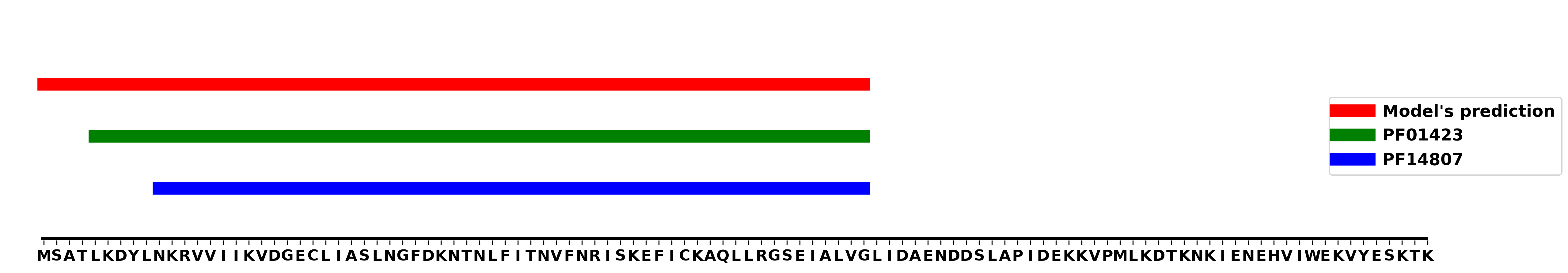}} \hfill
\subfloat[]{\label{fig:int_b}\includegraphics[width=\linewidth]{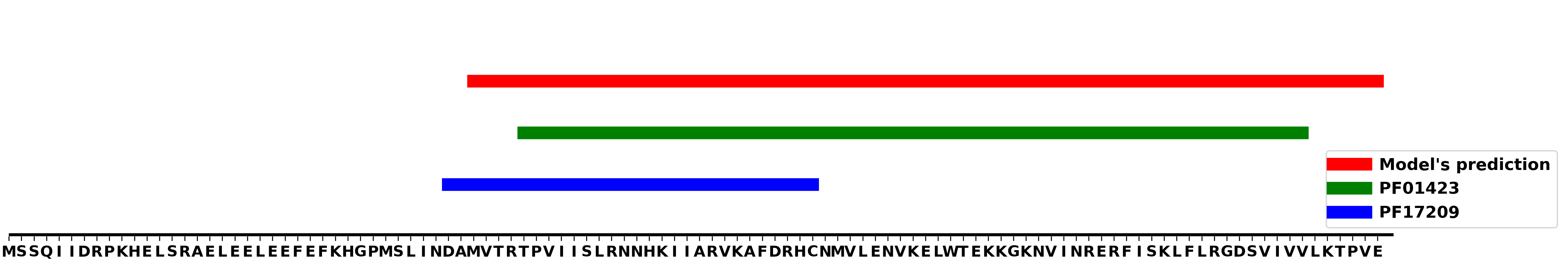}} \hfill
\subfloat[]{\label{fig:int_c}\includegraphics[width=\linewidth]{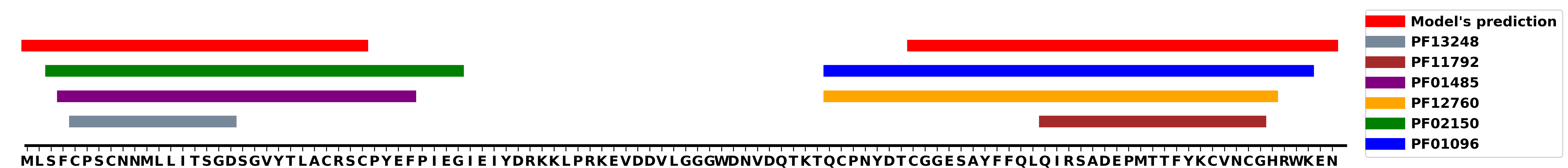}}
\caption{Visualization of motifs and the activated amino acids for three proteins (a) LSM8 (b) SMD2 and (c) RPC11. The legend includes the motif identifier from Pfam database. The alphabet on the x-axis is the amino acid at the respective position. The line segments corresponds to the part of sequence that is model's prediction (red) or motifs (other color). The selected subset of amino acid sequences aligns with motifs from Pfam motif library.}
\label{fig:interpretation}
\end{figure}

In addition, we visualize the activated amino acids and the motifs from Pfam motif library~\cite{finn2014pfam} from GenomeNet~\cite{kanehisa1997linking} in Figure~\ref{fig:interpretation}. For this experiment, we select three proteins: LSM8, SMD2, and RPC11 from the yeast dataset with motifs in different parts of the sequences. Figure~\ref{fig:ablation} and Table~\ref{motif_alignment} shows that the model trained with fusedmax achieves similar performance to sparsemax when $\mathbf{d} \geq 256$ but gains better alignment with the motifs from Pfam database. So, we train the model with fusedmax and obtain the gate values for each amino acid in these sequences. Red lines in each sub-figure of Figure~\ref{fig:interpretation} are the important regions identified by our model.

In particular, LSM8 with the sequence of length 109 has two motifs: PF01423, LSM domain at the position from 4 to 65, and PF14807, adaptin AP4 complex epsilon appendage platform at the position from 9 to 65 shown in Figure~\ref{fig:int_a}. The learned gate value corresponds to the subset of amino acids from position 1 to 65 and aligns with motifs. Similarly, the selected parts of sequences for SMD2 and RPC11 aligns with their motifs even though the motif lies in different parts of sequences. The quantitative and qualitative evaluation of gate vectors shows that our model successfully identifies important amino acids in the sequence for PPI prediction.

\section{Conclusion}
\label{section:conclusion}
We present a novel deep neural network to model and predict PPIs from variable-length protein sequences. Our proposed framework adopts a recurrent neural network to capture contextualized and sequential information from amino acid sequences. By incorporating a structured and sparse gating mechanism into the sequence encoder, our model successfully selects the important residues on the sequence to learn the sequence representation. Furthermore, our novel approach of encoding sequences to their representation makes the model efficient and scalable to a large number of interactions. Extensive experimental evaluations on various up-to-date datasets show its promising performance on binary PPI prediction task. Various case studies demonstrate the ability of our model to provide biological insights to interpret the predictions.

\section*{Acknowledgment}
This work was supported by the National Science Foundation [NSF-1062422 to A.H.], [NSF-1850492 to R.L.] and the National Institutes of Health [GM116102 to F.C.] 




%
\bibliographystyle{ieeetran}
\bibliography{IEEEfull}

\end{document}